\documentclass{article}
\usepackage{iclr2026_conference,times}

\usepackage{amsmath,amsfonts,bm}

\def\eqref#1{equation~\ref{#1}}

\def\1{\bm{1}}

\DeclareMathAlphabet{\mathsfit}{\encodingdefault}{\sfdefault}{m}{sl}
\SetMathAlphabet{\mathsfit}{bold}{\encodingdefault}{\sfdefault}{bx}{n}

\NewDocumentCommand{\yun}
{ mO{} }{\textcolor{green}{\textsuperscript{\textit{yun}}\textsf{\textbf{\small[#1]}}}}

\NewDocumentCommand{\zhuo}
{ mO{} }{\textcolor{red}{\textsuperscript{\textit{zhuo}}\textsf{\textbf{\small[#1]}}}}

\NewDocumentCommand{\yafu}
{ mO{} }{\textcolor{cyan}{\textsuperscript{\textit{yafu}}\textsf{\textbf{\small[#1]}}}}

\usepackage{hyperref}
\usepackage{url}
\usepackage{booktabs}
\usepackage{etoolbox}
\usepackage{amsfonts}
\usepackage{nicefrac}
\usepackage{microtype}
\usepackage{xcolor}
\usepackage{colortbl}
\usepackage{amsmath}
\usepackage{graphicx}
\usepackage{float}
\usepackage{wrapfig}
\usepackage{fvextra}
\usepackage{enumitem}
\usepackage{caption}
\usepackage{subcaption}
\usepackage[most]{tcolorbox}
\usepackage{tabularx}
\usepackage{amsthm}
\usepackage{bm}
\usepackage{multirow}
\usepackage[section]{placeins}
\tcbuselibrary{breakable}
\usepackage[capitalize,noabbrev]{cleveref}
\usepackage{titletoc}
\usepackage{tikz}

\definecolor{mintblue}{RGB}{210,235,250}
\definecolor{mintframe}{RGB}{120,180,220} 
\definecolor{minttitle}{RGB}{100,150,200} 
\definecolor{minttext}{RGB}{50,80,120}    

\definecolor{runzhemilk}{RGB}{255,235,245} 
\definecolor{roseframe}{RGB}{230,120,150}  
\definecolor{runzhecotton}{RGB}{255,170,200}    

\newtcolorbox{promptbox}[1]{
  enhanced,
  breakable,
  colback= runzhemilk!30!white,   
  colframe=roseframe,                
  colbacktitle= runzhecotton!66!white, 
  coltitle=white!33,
  title=\textbf{#1},
  fonttitle=\bfseries,
  sharp corners=south, 
  borderline={0.8pt}{0pt}{roseframe},
  boxrule=0.8pt,
  arc=6pt, 
  left=6pt, right=6pt, top=6pt, bottom=6pt,
  before skip=10pt, after skip=10pt,
  drop shadow=black!12,      
}

\newtcolorbox{casebox}[1]{
enhanced,
breakable,
colback=mintblue!40!white,
colframe=mintframe,
colbacktitle=minttitle!70!white,
coltitle=white,
title=\textbf{#1},
fonttitle=\bfseries,
sharp corners=south, 
borderline={0.8pt}{0pt}{minttitle},
boxrule=0.8pt,
arc=6pt, 
left=6pt, right=6pt, top=6pt, bottom=6pt,
before skip=10pt, after skip=10pt,
drop shadow=black!15, 
}

\newtcolorbox{takeawaysbox}{
enhanced,
breakable,
colback=mintblue!40!white,
colframe=mintframe,
colbacktitle=minttitle!70!white,
coltitle=white,
title=\textbf{Key Takeaways},
fonttitle=\bfseries,
sharp corners=south, 
borderline={0.8pt}{0pt}{minttitle},
boxrule=0.8pt,
arc=6pt, 
left=6pt, right=6pt, top=6pt, bottom=6pt,
before skip=10pt, after skip=10pt,
drop shadow=black!15, 
}

\makeatletter
\newcommand{\DrawLine}{%
  \begin{tikzpicture}
  \path[use as bounding box] (0,0) -- (\linewidth,0);
  \draw[color=minttitle!70!white,dashed,dash phase=1.5pt]
        (0-\kvtcb@leftlower-\kvtcb@boxsep,0)--
        (\linewidth+\kvtcb@rightlower+\kvtcb@boxsep,0);
  \end{tikzpicture}%
  }
\makeatother

\definecolor{darkblue}{rgb}{0, 0, 0.5}
\definecolor{citeblue}{rgb}{0.35, 0.62, 0.88}
\definecolor{methodrowgreen}{RGB}{235,248,239}
\hypersetup{colorlinks=true, citecolor=citeblue, urlcolor=citeblue, linkcolor=darkblue}

\newcommand{\methodname}{\ensuremath{\mathrm{SP}^{3}\mathrm{O}}}
\newcommand{\methodnamebold}{\ensuremath{\bm{\mathrm{SP}^{3}\mathrm{O}}}}

\title{Rethinking Critic Learning in PPO: Understanding and Mitigating Value Flattening}

\author{
\begin{minipage}{\textwidth}
\vspace*{0.6em}
\centering
\normalfont
\textbf{Yizhuo Li}\textsuperscript{1,2 *}\hspace{1em}
\textbf{Jianhao Yan}\textsuperscript{3 *}\hspace{1em}
\textbf{Yun Luo}\textsuperscript{2\textdagger,$\ddagger$}\hspace{1em}
\textbf{Zhi Wang}\textsuperscript{4}\hspace{1em}
\textbf{Futing Wang}\textsuperscript{2}\hspace{1em}
\textbf{Rong-Xi Tan}\textsuperscript{2,4}\hspace{1em}
\textbf{Kanghui Tian}\textsuperscript{2}\hspace{1em}
\textbf{Ganqu Cui}\textsuperscript{2}\hspace{1em}
\textbf{Ning Ding}\textsuperscript{5}\hspace{1em}
\textbf{Peilin Zhao}\textsuperscript{1$\ddagger$}\hspace{1em}
\textbf{Yafu Li}\textsuperscript{2,6$\ddagger$}\hspace{1em}
\textbf{Yu Cheng}\textsuperscript{7$\ddagger$}
\\[0.6em]
{\small
\textsuperscript{1} Shanghai Jiao Tong University \quad
\textsuperscript{2} Shanghai AI Laboratory \quad
\textsuperscript{3} Westlake University \quad
\textsuperscript{4} Nanjing University \quad \\
\textsuperscript{5} Tsinghua University \quad
\textsuperscript{6} The Chinese University of Hong Kong \quad
\textsuperscript{7} Nanyang Technological University
\\[0.3em]
\textsuperscript{*}Equal Contribution,$\quad$\textsuperscript{\textdagger}Project Lead,$\quad$ \textsuperscript{$\ddagger$}Corresponding authors
}
\end{minipage}
}

\iclrfinalcopy

\begin{document}

\maketitle

\begingroup
\renewcommand{\thefootnote}{}
\footnotetext{Code: \url{https://github.com/Dodojordi/SP3O}}
\endgroup

\vspace{-4mm}
\begin{abstract}

In reinforcement learning for large language models, Proximal Policy Optimization (PPO) commonly uses a critic to estimate state values and reduce the variance of policy updates. 
However, we uncover a systematic failure mode in PPO critics, which we call \textbf{Value Flattening}:
state values, estimated from multiple Monte Carlo continuations, change sharply across
intermediate states while critic predictions remain comparatively flat.
We further observe this phenomenon in a controlled FrozenLake environment and find that it becomes more pronounced as the state space grows.
Our theoretical and empirical analyses relate Value Flattening to an implicit variance penalty in the
critic loss and redundant updates from temporally correlated states with
similar gradients. Motivated by these findings, we introduce SParse Proximal Policy Optimization (\methodname), which applies
the value loss to only a few well-separated states in each response to mitigate both effects. Experiments on Qwen3-Base
show that
\methodname{} with only three states supervised per response can mitigate Value Flattening
and consistently improve the learned policy across model sizes and evaluation
suites. Together, our results identify Value Flattening as an important yet
overlooked failure mode of critic learning in standard PPO and show that a simple sparse supervision strategy can mitigate it.
\end{abstract}

\section{Introduction}
\label{sec:introduction}

Reinforcement learning enables large language models to tackle increasingly
challenging reasoning tasks that unfold over long
trajectories~\citep{k3,su01,chen2026omibench}.
Such long-horizon tasks pose a fundamental challenge for policy optimization, which requires both fine-grained credit assignment over intermediate decisions and informative advantage estimates at each generation step \citep{hou2026sao,kazemnejad2025vineppo,guo2025segmentpo,wang2026sppo,gong2026sae}.
Proximal Policy Optimization (PPO) offers a natural framework for meeting these
demands by using a critic to estimate state values throughout each trajectory
and construct token-level advantages~\citep{schulman2017ppo,schulman2015gae,yuan2025vcppo,yue2025vapo,qi2026bpco}.
This, in turn, places a key requirement on the critic: it must capture meaningful changes in expected success (i.e., the state value) across the trajectory.

\begin{figure}[t]
  \centering
  \includegraphics[width=\linewidth]{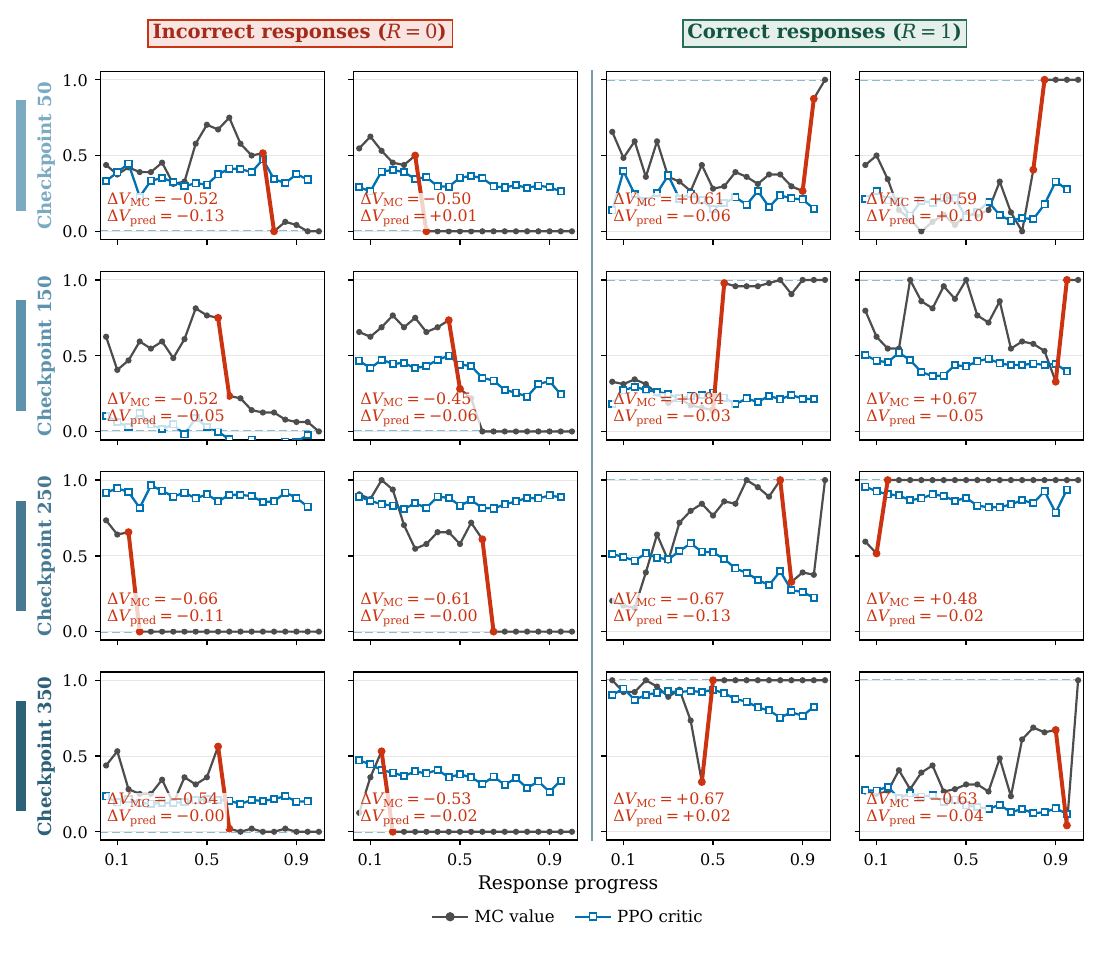}
  \captionsetup{belowskip=-14pt}
  \caption{Value Flattening across training in a Qwen3-4B-Base PPO run on
  DAPO-Math-17k. Each panel is a
  distinct correct or incorrect response selected for large local MC value changes. Across checkpoints, the corresponding PPO critic
  profiles remain comparatively flat, showing weak within-response
  resolution.}
  \label{fig:value-coarsening-cases-full}
  \vspace{-1mm}
\end{figure}

However, we find that the critic in standard PPO fails to capture substantial changes in state values across steps within a response.
We refer to this phenomenon as \textbf{Value Flattening}.
For each intermediate state, we estimate its state value by averaging terminal rewards from multiple independent continuations sampled from the same policy, obtaining Monte Carlo estimates of state values (MC values ~\citep{kazemnejad2025vineppo}). We use these estimates as diagnostic references for evaluating critic predictions
along individual trajectories.
\Cref{fig:value-coarsening-cases-full} illustrates this mismatch by comparing critic predictions with MC values.
Across these trajectories, MC values often exhibit sharp local transitions, whereas critic predictions remain comparatively flat.
In some cases, the critic prediction changes in the
opposite direction from the MC values.
These examples also show that critic predictions remain relatively insensitive to local variation in MC values across training checkpoints and in both correct
and incorrect responses, indicating that value flattening is a property of the
learned critic rather than a trajectory-specific effect.

{To further characterize Value Flattening, we study a stochastic FrozenLake environment, where we vary the maze size
while keeping the training configuration fixed. This setting allows us to directly compare critic predictions against state values. We observe that as the maze
size increases, critic predictions become progressively smoother and
less accurate (\Cref{fig:value-coarsening-diagnostics}b,c).}
Taken together, our observations in LLMs and FrozenLake suggest that Value Flattening is a systematic critic failure mode that may become more pronounced as the state space grows, motivating us to investigate its underlying causes.

To understand these phenomena, we analyze the critic training objective and the temporal dependence in its supervision, identifying two factors that can contribute to Value Flattening (\Cref{sec:value-coarsening-analysis}):
\textbf{(1) Implicit Variance Penalty:} Under the common setup in PPO training
with terminal-only rewards~\citep{yuan2025vcppo,yue2025vapo,
hu2025openreasonerzero,qi2026bpco}, the critic's mean squared error (MSE) loss
is applied at every token position in a response.
Our loss decomposition shows that minimizing this
objective
directly penalizes value variation across all token positions within a response,
pushing the predictions toward a flatter value profile.
\textbf{(2) Redundant Updates from Temporal Correlation:} LLM states consist of the tokens generated so far, so neighboring states differ by only one token and are highly temporally correlated. We find that nearby states have similar
representations and gradients in the critic, with gradient similarity decreasing as the distance between token positions increases. Dense token-level supervision therefore aggregates many similar updates from neighboring states, which can make their predicted values more similar.

Motivated by these findings, we propose SParse Proximal Policy Optimization (\methodname{}).
Supervising fewer, more widely separated positions restricts the implicit
variance penalty to their predicted values and is designed to reduce redundant
updates from temporally correlated states. Experiments show that \methodname{}
mitigates Value Flattening and consistently improves actor performance across
Qwen3-4B-Base and Qwen3-8B-Base on mathematical and out-of-distribution
reasoning benchmarks.
Our contributions are threefold:
\begin{itemize}[leftmargin=*,nosep]
  \item We identify Value Flattening, a systematic mismatch in which state values estimated from multiple Monte Carlo continuations can change sharply within individual responses while PPO critic predictions remain comparatively flat.
  \item We relate Value Flattening to two factors: the implicit variance penalty in the critic's mean squared error (MSE) loss, which penalizes value differences across token positions, and redundant updates from temporally correlated states with similar gradients.
  \item We introduce \methodname{}, which supervises the critic at a few well-separated states to mitigate the effects of the implicit variance penalty and redundant neighboring updates. \methodname{} mitigates Value Flattening and improves policy performance across model scales and evaluation suites.
  
\end{itemize}

\section{Related Work}
\label{sec:related-work}

\subsection{Critic-Free and Critic-Based RL}
Critic-free methods such as RLOO~\citep{ahmadian2024rloo}, GRPO~\citep{shao2024deepseekmath}, and DAPO~\citep{yu2025dapo} avoid training a value model, but their advantages are derived from complete responses and offer little direct distinction among states within the same response. Recent work such as VIMPO derives an implicit value function to recover finer-grained credit without a separate critic~\citep{kang2026vimpo}.
Critic-based methods such as PPO instead learn a value function and use its predictions at each generation step to construct token-level advantages~\citep{schulman2017ppo}.  Building on PPO, recent methods improve critic initialization and optimization for reasoning or adapt critic-based learning to asynchronous and structured trajectories~\citep{yuan2025vcppo,yue2025vapo,hou2026sao,li2026compactionrl,he2026simpleopdsimpletokenizeragnosticonpolicy,chen2025p1masteringphysicsolympiads,luo2026p1vlbridgingvisualperception,qi2026bpco}, but provide little analysis of whether their critic predictions capture value changes within individual responses. We study this behavior and identify Value Flattening: PPO critics retain differences across responses but fail to track value changes among states within the same response.

\subsection{Fine-Grained Credit Assignment and Value Estimation}
Prior work obtains intermediate credit through outcomes organized by segments or trees and process supervision~\citep{guo2025segmentpo,hou2025treerl,tran2025tempo,ielanskyi2026rredcot,lightman2023verify}. These signals provide finer feedback but do not directly estimate the policy-conditioned state value: the expected terminal return when the current policy continues from an intermediate state. VinePPO estimates this quantity with auxiliary continuations and improves credit assignment, but fine-grained estimation requires additional rollouts from each evaluated state and can become costly as responses lengthen~\citep{kazemnejad2025vineppo,wang2026sppo,gong2026sae,shan2026genac}. We build on this line of work by evaluating both response-level discrimination and the ability of PPO critics to track policy-conditioned state-value changes within individual responses across training stages and supervision densities.

\section{Preliminaries}
\label{sec:preliminary}

\subsection{Proximal Policy Optimization}
\label{sec:preliminary-ppo}

Given a prompt $x$, a policy $\pi_\theta$ generates a response
$y=(y_1,\ldots,y_T)$ and thereby a trajectory $\tau$, where the state and action at step $t$ are
$s_t=(x,y_{<t})$ and $a_t=y_t$.  The return from $s_t$ is
$G_t=\sum_{k=t}^{T}\gamma^{k-t}r_k$.
The per-step reward $r_t$ may combine a task reward with additional shaping
terms, such as a KL penalty.  PPO estimates advantages from trajectories
sampled by $\pi_{\theta_{\mathrm{old}}}$ and maximizes the clipped surrogate
objective~\citep{schulman2017ppo}
\begin{equation}
  \mathcal{L}_{\mathrm{PG}}(\theta)
  = \mathbb{E}_{t}\!\left[
      \min\!\left(
        \rho_t(\theta)\widehat A_t,
        \operatorname{clip}\!\left(\rho_t(\theta),1-\epsilon,1+\epsilon\right)
        \widehat A_t
      \right)
    \right],
  \label{eq:ppo-policy-objective}
\end{equation}
where
$\rho_t(\theta)=\pi_\theta(a_t\mid s_t)/
\pi_{\theta_{\mathrm{old}}}(a_t\mid s_t)$ and $\widehat A_t$ is the estimated
advantage.

PPO uses a critic $V_\phi(s_t)$ to approximate the policy-conditioned value
$V^\pi(s_t)=\mathbb{E}_\pi[G_t\mid s_t]$.  Generalized advantage estimation
(GAE) constructs the actor advantage $\widehat A_t$ and the critic regression
target $\widehat G_t$ as~\citep{schulman2015gae}:
\begin{equation}
  \resizebox{\columnwidth}{!}{$\displaystyle
  \delta_t=r_t+\gamma V_{\phi_{\mathrm{old}}}(s_{t+1})-V_{\phi_{\mathrm{old}}}(s_t),
  \qquad \widehat A_t=\sum_{\ell=0}^{T-t}(\gamma\lambda)^\ell\delta_{t+\ell},
  \qquad \widehat G_t=\widehat A_t+V_{\phi_{\mathrm{old}}}(s_t)$},
\label{eq:gae}
\end{equation}
where $V_{\phi_{\mathrm{old}}}(s_{T+1})=0$.  For a rollout batch $\mathcal B$, the
standard critic update is:
\begin{equation}
  \mathcal{L}_{V}(\phi)
  = \frac{1}{\sum_{\tau\in\mathcal{B}}T_\tau}
    \sum_{\tau\in\mathcal{B}}\sum_{t=1}^{T_\tau}
    \left(V_\phi(s_t)-\widehat G_t\right)^2,
  \qquad
  \phi \leftarrow \phi-\eta_V\nabla_\phi\mathcal{L}_{V}(\phi).
  \label{eq:critic-update}
\end{equation}
For the experiments in this work, the KL coefficient is zero and the task
reward is terminal. Let $R(\tau)$ denote the realized terminal task reward of
trajectory $\tau$, which is binary in our experiments. Thus, $r_t=0$ for
$t<T$ and $r_T=R(\tau)$; with $\gamma=\lambda=1$,
\[
  \widehat G_t=G_t=R(\tau), \qquad t=1,\ldots,T.
\]
This equality holds for the targets observed along a sampled trajectory; it
does not imply that the policy-conditioned value $V^\pi(s_t)$ is constant
within that trajectory. Standard critic training therefore repeats one
response-level outcome at every state, which is the supervision structure
studied in this work.

\subsection{Monte Carlo Estimation of State Value}
\label{sec:preliminary-mc}

A sampled return $G_t$ is a single-rollout Monte Carlo estimate of
$V^\pi(s_t)$. We obtain a lower-variance diagnostic estimate by independently
sampling $K$ continuations $\tau_t^{(1)},\ldots,\tau_t^{(K)}$ from the same
policy conditioned on a fixed state $s_t$:
\begin{equation}
  \widehat V_{\mathrm{MC}}^{\pi}(s_t)
  = \frac{1}{K}\sum_{k=1}^{K}G_t^{(k)},
  \label{eq:mc-state-value}
\end{equation}
where $\tau_t^{(k)}\sim\pi(\cdot\mid s_t)$ and
$G_t^{(k)}=R(\tau_t^{(k)})$ is its realized return in our terminal-reward
setting. This sample average estimates the theoretical state value
$V^\pi(s_t)=\mathbb E_\pi[G_t\mid s_t]$. For the terminal binary-reward setting
used here, it is the empirical success rate of the sampled continuations. We
use it as a diagnostic reference to distinguish fitting the single-rollout
target $\widehat G_t$ from agreement with the policy-conditioned state value.

\section{Understanding and Mitigating Value Flattening}
\label{sec:value-coarsening}

We first show that Value Flattening occurs in both LLM reasoning and a
controlled Markov decision process, and that it becomes more pronounced as the
state space grows. We then explain why the critic loss favors similar values
within a response and why dense supervision can reinforce this effect. Shared
terminal-return targets create an implicit variance penalty, while
temporally correlated states produce redundant updates under dense supervision. This
analysis motivates sparse critic supervision. Unless otherwise stated, the LLM
analysis uses Qwen3-4B-Base trained on DAPO-Math-17k. Full training and
figure-specific settings are provided in Appendix
\ref{app:experimental-details}.

\begin{figure}[tp]
  \centering
  \includegraphics[width=\textwidth]{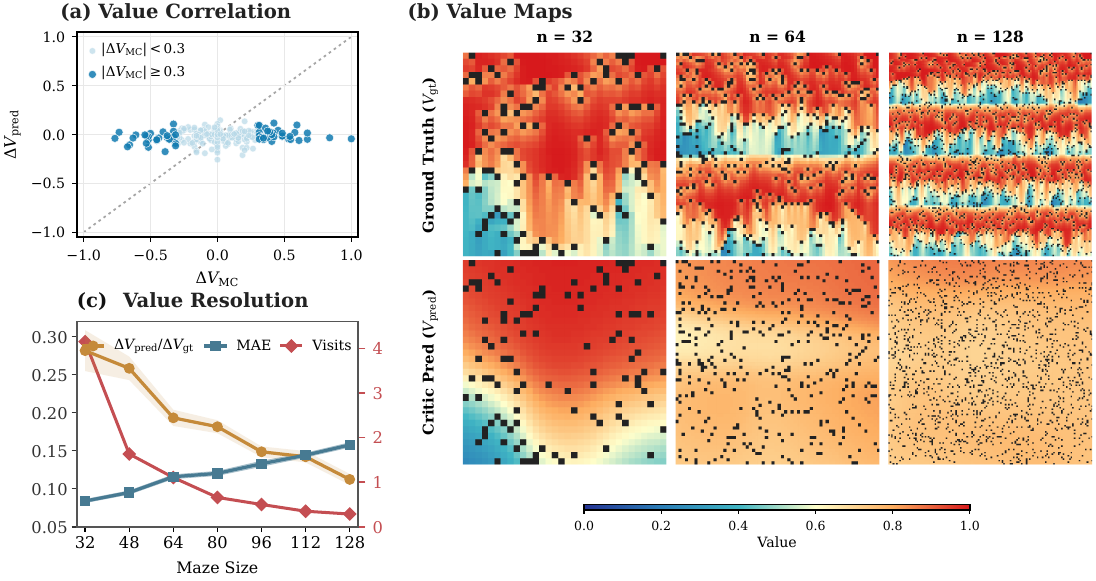}
  \captionsetup{font=small}
  \caption{Value Flattening in LLM reasoning and stochastic FrozenLake.
  \textbf{(a)} Adjacent MC and critic value changes: predicted changes remain
  concentrated near zero even when MC values change sharply.
  \textbf{(b)} Ground-truth and critic value maps as the FrozenLake maze size
  $n$ increases under an otherwise fixed training configuration.
  \textbf{(c)} The corresponding value-resolution statistics.}
  \label{fig:value-coarsening-diagnostics}
\end{figure}

\subsection{Value Flattening in Critic Learning}
\label{sec:value-coarsening-measurement}

\paragraph{Value Flattening in LLM Reasoning.}
We characterize Value Flattening directly through critic predictions. Across
training checkpoints and for both correct and incorrect responses, MC values
can change sharply between adjacent reasoning states while the corresponding
critic profiles remain comparatively flat
(\cref{fig:value-coarsening-cases-full}). The adjacent-change analysis in
\cref{fig:value-coarsening-diagnostics}a makes this mismatch explicit: MC value
changes span a broad range, whereas critic changes remain concentrated near
zero instead of following the diagonal. The critic therefore often fails to
capture both the magnitude and the direction of local value changes.
Consequently, reasoning states with substantially different MC values can
receive similar critic predictions within the same response. Appendix
\ref{app:theoretical-insights} formalizes this mismatch through an error
decomposition.

\paragraph{Value Flattening with State-Space Growth.}
To examine whether Value Flattening extends beyond LLMs and how it scales with state-space size, we study stochastic FrozenLake~\citep{brockman2016openai}, where an agent navigates from a start
state to a goal while avoiding holes. Each episode has a binary return, equal to
$1$ upon reaching the goal and $0$ otherwise, so the state value
of a grid cell is the probability of eventually reaching the goal from that
state. This setting provides a controlled experiment in which we hold the
training configuration fixed and increase the maze size $n$ to expand the
state space. As shown in \Cref{fig:value-coarsening-diagnostics}b, critic predictions across grid cells become progressively smoother as the maze grows. The
local-contrast and error metrics in panel (c) likewise show smoother value
predictions and weaker agreement with the ground truth. Thus, Value Flattening
becomes more pronounced as the state space grows.

\subsection{Why Does Value Flattening Occur?}
\label{sec:value-coarsening-analysis}
By analyzing the critic objective and the temporal  dependence among supervised states, we identify two factors contributing to Value Flattening: an implicit variance penalty and redundant critic updates.
\paragraph{Implicit Variance Penalty.}
Under terminal-only rewards with $\gamma=\lambda=1$, standard PPO uses the same
sampled terminal return as the target at every state in a response. Its critic
loss therefore contains an implicit variance penalty. For a response of length
$T$, let $v_t=V_\phi(s_t)$ and $\bar v=T^{-1}\sum_{t=1}^{T}v_t$. Then
\begin{equation}
  \frac{1}{T}\sum_{t=1}^{T}(v_t-R)^2
  = (\bar v-R)^2 + \frac{1}{T}\sum_{t=1}^{T}(v_t-\bar v)^2.
  \label{eq:single-response-variance-decomposition}
\end{equation}
The first term fits the mean prediction to the sampled response outcome. The
second term is the empirical variance of the predictions and directly penalizes
their variation within that response. This variance penalty arises because
every position uses the same terminal target. Under dense
supervision, this variance is computed over all $T$ positions, so every token
prediction is directly included in the penalty. The corresponding batch-level
loss and parameter-gradient decompositions are given in Appendix
\ref{app:theoretical-insights}.

\begin{figure*}[t]
  \centering
  \captionsetup{skip=4pt}
  \includegraphics[width=\textwidth]{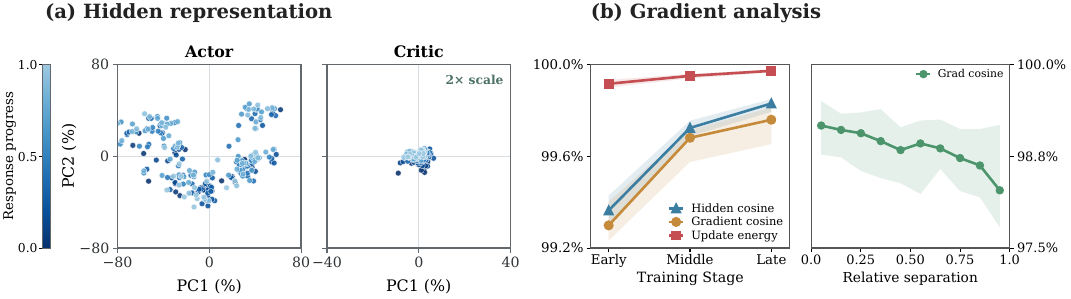}
  \caption{Representation and gradient analysis in Qwen3-4B-Base PPO critic
  supervision. \textbf{(a)} Actor and critic hidden representations for
  the same response, centered and normalized by mean hidden norm; the critic
  view uses a 2$\times$ scale. \textbf{(b)} Hidden-representation alignment,
  gradient alignment, and update energy remain high across PPO training stages (left), while the full critic
  gradient cosine decreases with relative token-position separation (right).}
  \label{fig:value-coarsening-scaling}
\end{figure*}

\paragraph{Redundant Updates from Temporal Correlation.}
Classical RL often operates on compact Markov states that summarize the
information needed for future decisions. In LLM post-training, by contrast, a
state contains the tokens generated so far, so adjacent states differ by only
one token and overlap in nearly their entire input. Because adjacent states are
temporally correlated and share most of their tokens, supervising all of them provides less diverse training signals
than their number suggests~\citep{mnih2015human}. This similarity is also reflected in the
critic hidden representations, whose trajectory in
\cref{fig:value-coarsening-scaling}a occupies a compact region.

Representation similarity connects directly to update similarity. Consider a
value head $V_\phi(s_t)=w^\top h_t$, where $h_t$ is the critic representation.
The gradient from position $t$ is
$g^{(w)}_t=2\bigl(V_\phi(s_t)-R\bigr)h_t$. Thus, positions with similar
representations and residuals of the same sign produce aligned gradients. The
measurements in \cref{fig:value-coarsening-scaling}b show that hidden-state
alignment, gradient alignment, and update energy remain high throughout training. Within
each response, gradient similarity decreases as the distance between their token
positions increases.
These results indicate that dense supervision produces many redundant updates
from neighboring states.

\subsection{\methodname{}: Sparse Critic Supervision}
\label{sec:value-coarsening-method}
Sparse supervision is designed to mitigate the two problems above through
selection and spacing. Selecting fewer positions restricts the per-response
variance penalty to those predictions instead of applying it at every token
position. Spacing
the selected positions apart is designed to reduce the accumulation of aligned
gradients from nearby token positions whose states share most of their tokens.

We instantiate this intervention as SParse Proximal Policy Optimization
($\methodname{}$).  The actor objective, rollout procedure, and return targets
remain unchanged; only the states receiving the critic loss are changed.
$\methodname{}$ applies the value loss only at a small
set of well-separated states.  The critic still produces values at every
generation state.  Let $\mathcal I(\tau)$ denote the supervised states in
trajectory $\tau$.  The sparse critic objective is
\begin{equation}
  \mathcal L_V^{\methodname{}}(\phi)
  = \frac{1}{\sum_{\tau\in\mathcal B}|\mathcal I(\tau)|}
    \sum_{\tau\in\mathcal B}\sum_{t\in\mathcal I(\tau)}
    \left(V_\phi(s_t)-\widehat G_t\right)^2.
  \label{eq:sparse-critic-objective}
\end{equation}

\begin{figure}[t]
  \centering
  \includegraphics[width=\textwidth]{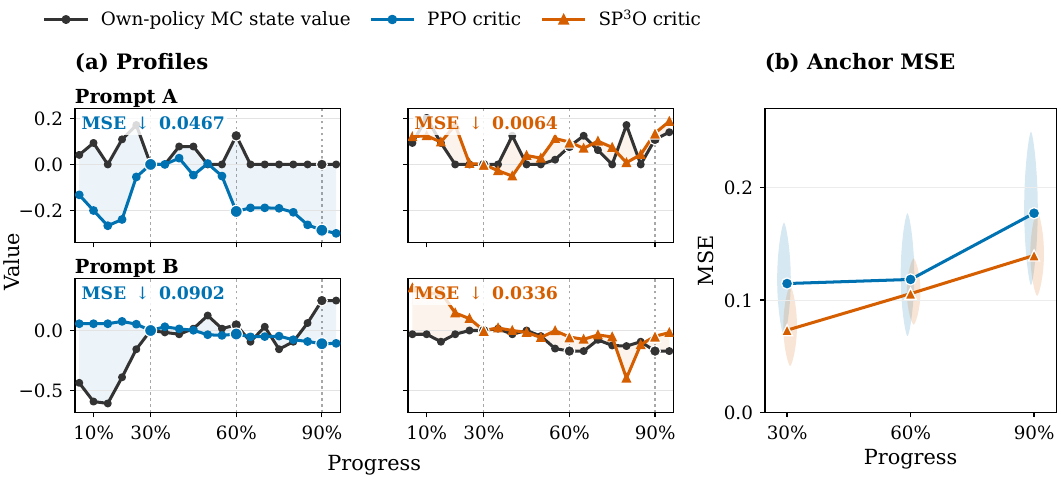}
  \caption{Effects on critic prediction in Qwen3-4B-Base. \textbf{(a)} Prompt-matched critic and
  own-policy MC value profiles. \textbf{(b)} Response-centered value
  error.}
  \label{fig:value-resolution-recovery-cases}
\end{figure}

\paragraph{Critic Prediction Effects.}
We compare both critics with policy-conditioned MC values obtained from
repeated continuations at intermediate states. As shown in
\cref{fig:value-resolution-recovery-cases}(a), PPO produces relatively flat
value profiles and misses substantial local changes in MC values, whereas
\methodname{} more closely captures their direction and magnitude. Its profile
MSE is lower on both selected prompts. The aggregate comparison in (b) confirms
this trend, with \methodname{} reducing MSE by \(36\%\), \(11\%\), and \(21\%\)
at \(30\%\), \(60\%\), and \(90\%\) response progress, respectively. These
results indicate that sparse supervision better preserves within-response value
variation and improves within-response value resolution.

\begin{figure*}[t]
  \centering
  \captionsetup{skip=2pt}
  \includegraphics[width=\textwidth]{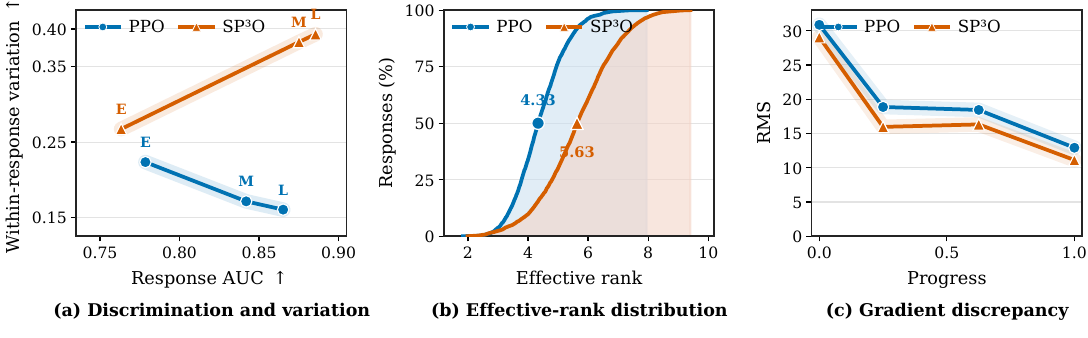}
  \caption{Effects on critic optimization in Qwen3-4B-Base. \textbf{(a)} Response AUC,
  computed from the mean critic prediction per response and its binary terminal
  outcome, versus the within-response fraction of critic-prediction variance.
  \textbf{(b)} Empirical cumulative distribution of the effective rank of each
  response's hidden-state matrix, measuring the dimensional diversity of
  value-head inputs across intermediate states. \textbf{(c)} RMS across
  responses of the difference between value-head gradients computed using
  terminal-return and policy-conditioned MC targets. Full definitions and
  aggregation details are provided in Appendix~\ref{app:experimental-details}.}
  \label{fig:value-coarsening-structure}
  \vspace{-1.2\baselineskip}
\end{figure*}

\paragraph{Critic Optimization Effects.}
Figure \ref{fig:value-coarsening-structure} examines how sparse supervision affects critic representations and
optimization. Panel (a) compares response-level outcome discrimination with
within-response critic-prediction variation across the early (\textbf{E}), middle
(\textbf{M}), and late (\textbf{L}) training stages.
PPO gains discrimination while losing within-response variation, whereas
\methodname{} improves both. In panel (b), \methodname{} shifts the effective
rank of response-centered hidden states upward, increasing the median from
\(4.33\) to \(5.63\) and indicating richer representations of state differences.
Panel (c) shows lower RMS discrepancy between value-head gradients induced by
terminal-return targets and MC values throughout the response. Together, these
observations are consistent with sparse supervision preserving richer hidden
representations and reducing redundant or distorted critic updates, without
sacrificing response-level discrimination.

\paragraph{Actor Optimization Effects.}
The critic is intended to stabilize PPO updates by replacing raw outcome
signals with advantage estimates that reduce policy-gradient variance. Its
within-response resolution can therefore affect both the direction and the
stability of actor optimization. In the online runs, $\methodname{}$ has smaller
within-iteration update changes over most of training than PPO
(\cref{fig:online-training-and-update}a). This pattern is consistent with
smoother actor optimization while retaining the benefits of a learned critic.

\section{Experiments}
\label{sec:experiment}

\subsection{Experimental Setup}
\label{sec:experiment-setup}

\paragraph{Models and baselines.}
We evaluate the proposed sparse critic supervision on Qwen3-4B-Base and
Qwen3-8B-Base~\citep{yang2025qwen3}. We train both models on
DAPO-Math-17k~\citep{yu2025dapo}. For each model, we report the initial checkpoint as a
reference and compare standard PPO, GRPO, and PPO with sparse critic supervision
and explicit late-tail coverage.  We refer to the last variant as SParse
Proximal Policy Optimization (\methodname{}).  Unless otherwise stated,
\methodname{} supervises response-relative states at $0.3$, $0.6$, and $0.9$,
adding a $0.95$ state for responses of at least
6144 tokens.

\paragraph{Benchmarks and evaluation.}
We use the 24K evaluation setting reported in the result sheets.  The
mathematical-reasoning suite contains AIME24, AIME25, AIME26, and
AMC23~\citep{maaMathCompetitions}, MATH500~\citep{lightman2023verify},
Minerva~\citep{lewkowycz2022solving}, and OlympiadBench~\citep{he2024olympiadbench}.
The out-of-distribution
(OOD) suite contains ARC-C~\citep{clark2018arc},
MMLU-Pro~\citep{wang2024mmlupro}, GPQA~\citep{rein2024gpqa},
AGIEval-English~\citep{zhong2024agieval},
BigBenchHard~\citep{suzgun2023challenging}, and
ZebraLogic-Grid~\citep{wildEvalZebraLogic}. Mathematical scores
are averaged over 32 generations.  For OOD evaluation, ARC-C,
MMLU-Pro, GPQA, AGIEval-English, BigBenchHard, and ZebraLogic-Grid are all
averaged over four generations.  The latter three benchmarks are
scored using xVerify~\citep{chen2025xverify}.

\begin{table}[!t]
  \centering
  \footnotesize
  \setlength{\tabcolsep}{3.2pt}
  \resizebox{\textwidth}{!}{%
  \begin{tabular}{llcccccccc}
    \toprule
    \multirow{2}{*}{\textbf{Model}}
    & \multirow{2}{*}{\textbf{Method}}
    & \multicolumn{8}{c}{\textbf{Mathematical Reasoning (avg@32)}} \\
    \cmidrule(lr){3-10}
    & & \textbf{AIME24} & \textbf{AIME25} & \textbf{AIME26}
    & \textbf{AMC23} & \textbf{MATH500} & \textbf{Minerva}
    & \textbf{Olympiad} & \textbf{Avg.} \\
    \midrule
    \multirow{4}{*}{Qwen3-4B-Base}
    & Base        & 4.58  & 3.75  & 4.38  & 25.08 & 42.78 & 22.71 & 22.35 & 17.95 \\
    & PPO         & 17.50 & 19.90 & 14.69 & 61.80 & 70.15 & 42.82 & 36.35 & 37.60 \\
    & GRPO        & 17.19 & 16.19 & 10.10 & 63.83 & 78.21 & 45.71 & 43.62 & 39.26 \\
    \rowcolor{methodrowgreen}
    & \methodnamebold{} & \textbf{23.02} & \textbf{23.54} & \textbf{22.08}
                  & \textbf{69.92} & \textbf{83.79} & \textbf{47.93}
                  & \textbf{48.68} & \textbf{45.57} \\
    \midrule
    \multirow{4}{*}{Qwen3-8B-Base}
    & Base        & 7.40  & 7.92  & 5.94  & 37.66 & 54.08 & 24.69 & 27.36 & 23.58 \\
    & PPO         & 30.21 & 25.10 & 23.33 & 72.19 & 86.33 & 48.81 & 53.56 & 48.50 \\
    & GRPO        & 28.50 & 22.04 & 22.70 & 73.82 & 85.26 & \textbf{52.06} & 50.98 & 47.91 \\
    \rowcolor{methodrowgreen}
    & \methodnamebold{} & \textbf{33.91} & \textbf{28.02} & \textbf{27.39}
                  & \textbf{75.00} & \textbf{87.33} & 47.40
                  & \textbf{54.50} & \textbf{50.51} \\
    \bottomrule
  \end{tabular}%
  }
  \caption{In-domain mathematical-reasoning accuracy (\%).  Each score is
  averaged over 32 generations (avg@32), and the
  final column averages the seven listed tasks.  Our method is shaded in light
  green, and bold denotes the best result within each model block.}
  \label{tab:main-results}
\end{table}

\begin{table}[!t]
  \vspace*{-2pt}
  \centering
  \footnotesize
  \setlength{\tabcolsep}{4.2pt}
  \resizebox{\textwidth}{!}{%
  \begin{tabular}{llccccccc}
    \toprule
    \multirow{2}{*}{\textbf{Model}}
    & \multirow{2}{*}{\textbf{Method}}
    & \multicolumn{7}{c}{\textbf{General Reasoning (avg@4)}} \\
    \cmidrule(lr){3-9}
    & & \textbf{ARC-C} & \textbf{MMLU-Pro} & \textbf{GPQA}
    & \textbf{AGIEval}$^{\dagger}$ & \textbf{BBH}$^{\dagger}$
    & \textbf{ZebraLogic}$^{\dagger}$ & \textbf{Avg.} \\
    \midrule
    \multirow{4}{*}{Qwen3-4B-Base}
    & Base        & 34.98 & 16.17 & 14.02          & 27.93          & 21.11          & 1.35           & 19.26 \\
    & PPO         & 89.19 & 54.39 & 34.85          & 65.87          & 57.50          & 9.90           & 51.95 \\
    & GRPO        & 90.96 & 56.87 & \textbf{38.89} & 68.18          & 71.17          & 12.58          & 56.44 \\
    \rowcolor{methodrowgreen}
    & \methodnamebold{} & \textbf{91.02} & \textbf{61.75} & \textbf{38.89}
                  & \textbf{71.44} & \textbf{73.35} & \textbf{19.20} & \textbf{59.28} \\
    \midrule
    \multirow{4}{*}{Qwen3-8B-Base}
    & Base        & 61.82          & 36.22          & 27.02          & 48.59          & 50.71          & 4.90           & 38.21 \\
    & PPO         & \textbf{93.84} & 64.19          & 47.22          & 74.52          & 79.27          & 27.25          & 64.38 \\
    & GRPO        & 93.04          & \textbf{66.27} & 49.49          & 75.15          & 78.10          & 27.40          & 64.91 \\
    \rowcolor{methodrowgreen}
    & \methodnamebold{} & 93.13          & 65.83          & \textbf{49.94} & \textbf{76.95}
                  & \textbf{80.89} & \textbf{31.45} & \textbf{66.37} \\
    \bottomrule
  \end{tabular}%
  }
  \caption{Out-of-distribution evaluation accuracy (\%).  All scores are
  averaged over four generations (avg@4).  Datasets
  marked with $\dagger$ are scored by the \texttt{gpt-oss-120b} verifier
  through xverify.  The final column is the unweighted mean over all six tasks.
  Our method is shaded in light green, and bold denotes the best result within
  each model block.}
  \label{tab:ood-results}
  \vspace*{-3pt}
\end{table}

\subsection{Main Results}
\label{sec:experiment-main-results}

Tables \ref{tab:main-results} and \ref{tab:ood-results} show that
\methodname{} consistently outperforms standard PPO and GRPO across both model
sizes and evaluation suites. The gains over PPO reach 7.97 percentage points on
in-domain mathematical reasoning and 7.33 percentage points on
out-of-distribution reasoning, with positive improvements also observed for
Qwen3-8B-Base.
These results indicate that mitigating critic value flattening
through sparse supervision leads to better policy learning rather than merely
improving critic-side diagnostics.

\paragraph{Online learning dynamics.}
\Cref{fig:online-training-and-update} compares the training dynamics of PPO and
\methodname{} on Qwen3-4B-Base. As shown in (a), \methodname{} exhibits smaller
and less variable within-iteration actor updates, indicating more stable
training. It also maintains higher validation accuracy and rollout reward than
PPO after the early training stage, as shown in (b) and (c). Meanwhile,
\methodname{} produces longer responses, whereas response lengths under PPO increase more gradually and remain shorter, as shown in (d). Overall,
\methodname{} achieves stronger performance while
maintaining more stable policy updates.

\begingroup
\setlength{\intextsep}{4pt}
\begin{figure}[!htbp]
  \centering
  \captionsetup{skip=2pt}
  \includegraphics[width=\textwidth]{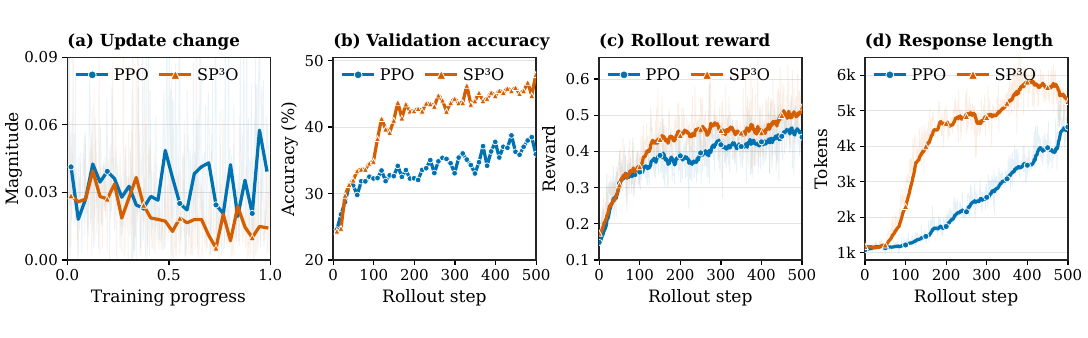}
  \caption{Online learning dynamics and within-iteration actor updates on
  Qwen3-4B-Base. \textbf{(a)} Within-iteration update change.
  \textbf{(b)} Validation accuracy. \textbf{(c)} Rollout reward.
  \textbf{(d)} Response length.}
  \label{fig:online-training-and-update}
\end{figure}
\endgroup

\FloatBarrier
\subsection{Sparse-Supervision Ablations}
\label{sec:sparse-supervision-ablations}
We further examine how the effectiveness of \methodname{} varies with supervision density, anchor placement, and late-tail coverage.

\noindent
\begin{minipage}[t]{0.52\linewidth}
\vspace{0pt}
\paragraph{Number of supervised states.}
We vary the number of supervised anchors $K$ during reasoning training with Qwen3-4B-Base to examine the effect of critic-supervision density (\cref{fig:anchor-count-performance-line}). Sparse configurations with $K\in\{3,4,8\}$ achieve higher mean training rewards than both denser configurations and standard PPO with token-level critic supervision. Performance is highest at $K=3$ and remains relatively stable through $K=8$, but drops substantially at $K=16$ and $K=64$, approaching the dense PPO baseline. Although the variance across runs is non-negligible, the overall trend indicates that increasing the number of supervised states does not improve reasoning performance. Instead, a small set of well-spaced anchors appears to provide sufficient coverage while limiting redundant critic updates.
\end{minipage}%
\hfill
\begin{minipage}[t]{0.44\linewidth}
\vspace{0pt}
  \centering
  \includegraphics[width=\linewidth]{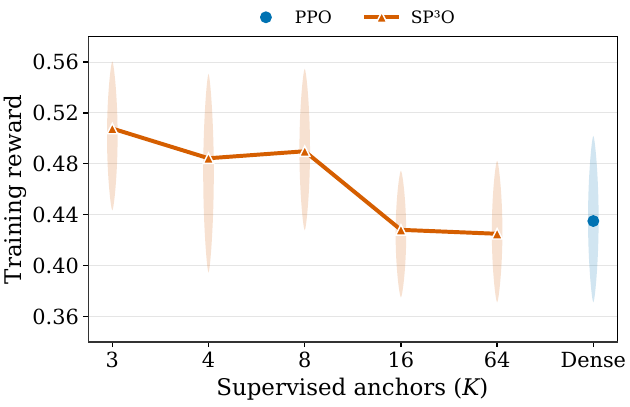}
  \captionsetup{skip=2pt}
  \captionof{figure}{Effect of critic-supervision density on performance.}
  \label{fig:anchor-count-performance-line}
\end{minipage}

\paragraph{Placement and late-tail coverage.}
\begin{wraptable}{r}{0.45\linewidth}
  \vspace{-0.8\baselineskip}
    \centering
  \footnotesize
  \setlength{\tabcolsep}{6pt}
  \begin{tabular*}{\linewidth}{@{\extracolsep{\fill}}lr@{}}
    \toprule
    \textbf{Main anchors} & \textbf{Acc. (\%)} \\
    \midrule
    PPO baseline  & 37.60 \\
    Random        & 36.59 \\
    $0.2/0.5/0.8$ & 44.65 \\
    $0.3/0.6/0.9$ & \textbf{45.57} \\
    \bottomrule
  \end{tabular*}
  \caption{Anchor-placement ablation on Qwen3-4B-Base ($K=3$).}
  \label{tab:anchor-ablation}

\end{wraptable}
At a fixed supervision density of $K=3$, both fixed-position placement schemes
outperform random placement, which performs worst (Table~\ref{tab:anchor-ablation}
and Figure~\ref{fig:anchor-position-ablation-curves}). This pattern is
consistent with temporal correlation playing a role: well-spaced anchors cover
the trajectory while avoiding repeated updates on nearby states that share most
of their history.

We further compare \methodname{} with a variant that removes the final tail
anchor while keeping all other supervised anchors unchanged. Adding the tail
anchor improves performance from \(44.10\) to \(45.57\) and reduces repetition
from \(18.33\) to \(1.12\) (Appendix Table~\ref{tab:anchor-ablation-tail}),
highlighting the importance of explicitly covering the response tail for policy
performance and generation stability.

\section{Conclusion}
\label{sec:conclusion}

We identify Value Flattening as a systematic failure mode of critics for large
language model reasoning, in which critic predictions fail to capture
policy-conditioned state value changes within individual responses. The same pattern
appears in controlled stochastic FrozenLake experiments and becomes more
pronounced as the state space grows. Our analyses relate this behavior to two
factors: an implicit variance penalty from dense token-level supervision and redundant
updates from temporally correlated states with similar gradients. Motivated
by these findings, \methodname{} applies the critic loss at a few well-separated
states to mitigate the effects of both factors. Experiments on Qwen3-4B-Base
and Qwen3-8B-Base show that \methodname{} mitigates Value Flattening and
improves actor performance across mathematical and out-of-distribution
reasoning benchmarks. These results establish Value Flattening as an important
yet overlooked problem in critic learning.

\section*{Acknowledgments}

This work was supported by the Shanghai Artificial Intelligence Laboratory. We are grateful to the authors and open-source communities whose work made this project possible.

\bibliography{paper}
\bibliographystyle{iclr2026_conference}
\appendix
\clearpage
\section{Appendix}
\label{app:appendix}

\subsection{Experimental Details and Diagnostic Metrics}
\label{app:experimental-details}

\begin{table}[!ht]
  \centering
  \footnotesize
  \setlength{\tabcolsep}{6pt}
  \begin{tabular}{lp{0.61\linewidth}}
    \toprule
    \textbf{Setting} & \textbf{Value} \\
    \midrule
    Training data & DAPO-Math-17k \\
    Rollout batch & $64\times8=512$ \\
    Actor update batch & $256/\mathrm{step}$; $2/\mathrm{rollout}$ \\
    Response length / temperature & $8{,}192/1.0$ \\
    Actor learning rate & $1\times10^{-6}$ \\
    Critic learning rate & $4\times10^{-6}$ \\
    Optimizer & Adam; $\beta=(0.9,0.98)$; $\mathrm{wd}=0.1$ \\
    PPO ratio clip / KL coefficient & $0.2/0$ \\
    Critic-only warm-up & $20$ batches \\
    \bottomrule
  \end{tabular}
  \caption{Key training hyperparameters for the Qwen3-4B-Base and
  Qwen3-8B-Base experiments.}
  \label{tab:training-hyperparameters}
\end{table}

\paragraph{Figure and table settings.}
In \cref{fig:anchor-count-performance-line}, the $K=4$ point uses anchors at
$0.3/0.5/0.7/0.9$. Points show mean performance, with bands indicating
variation.

\paragraph{FrozenLake settings.}
Each rollout terminates when the agent reaches the goal, falls into a hole, or
reaches the maximum episode length of 8192 steps. Reaching the goal yields a
return of $1$, while falling into a hole or reaching the episode limit yields a
return of $0$.

\paragraph{Policy-conditioned Monte Carlo estimation.}
For the fixed evaluation panel, we use 64 responses and evaluate each response
at 20 anchors: 19 intermediate states at relative positions
$0.05,0.10,\ldots,0.95$, plus the completed response at position $1.0$.
At each intermediate state, we hold the state fixed and independently sample
continuations $\tau_t^{(k)}\sim\pi(\cdot\mid s_t)$ from the same actor policy.
We start with 128 continuations and add batches of 64, up to a maximum of 256,
stopping earlier when the half-width of the 95\% Wilson confidence interval for
the empirical success rate is at most $0.04$. Thus, the
nonterminal anchors use $K\in\{128,192,256\}$. We estimate the
policy-conditioned state value by averaging the realized terminal rewards,
$\widehat V_{\mathrm{MC}}^\pi(s_t)=K^{-1}\sum_{k=1}^{K}G_t^{(k)}$. We use
temperature $1.0$, top-$p$ $0.95$, and at most $8192$ generated tokens per
continuation. The terminal anchor uses its observed terminal reward $R(\tau)$
directly rather than additional sampling.

\paragraph{Critic diagnostic metrics.}
For each response $i$, let $\Delta v_{i,t}$ denote the change in the predicted
value at evaluated state $t$. We define the \emph{update energy} in
\cref{fig:value-coarsening-scaling}b as
\begin{equation}
  E_i=
  \frac{T_i\overline{\Delta v}_i^{\,2}}
       {\sum_t\Delta v_{i,t}^{2}},
  \qquad
  \overline{\Delta v}_i=\frac{1}{T_i}\sum_t\Delta v_{i,t}.
  \label{eq:update-energy}
\end{equation}

Let $v_{i,t}=V_\phi(s_{i,t})$ be the critic prediction at token $t$ of response
$i$, $T_i$ its number of valid tokens, and $R_i\in\{0,1\}$ its terminal
outcome. \emph{Response AUC} is the ROC-AUC of $R_i$ scored by the mean
prediction $\bar v_i=T_i^{-1}\sum_t v_{i,t}$. The \emph{within-response
variation} ratio is
\begin{equation}
  \frac{\sum_i\sum_t(v_{i,t}-\bar v_i)^2}
       {\sum_i\sum_t(v_{i,t}-\bar v)^2},
  \qquad
  \bar v=\frac{\sum_i\sum_t v_{i,t}}{\sum_i T_i},
  \label{eq:within-response-variation}
\end{equation}
where lower values indicate flatter predictions within responses. Both metrics
are computed per rollout batch and summarized by stage medians in
\cref{fig:value-coarsening-structure}a. For panel (b), we unit-normalize the
value-head input states within each response, subtract their response mean, and
stack them into a matrix $H_i$. If $\lambda_j$ are the eigenvalues of
$H_iH_i^\top$ and $p_j=\lambda_j/\sum_k\lambda_k$, its \emph{effective rank}
is $r_{\mathrm{eff}}=\exp(-\sum_jp_j\log p_j)$; the panel plots its empirical
cumulative distribution across responses. For panel (c), let $h_{i,t}$ denote
the value-head input and $A_i$ the evaluated positions. The response-level mean
gradient difference under the squared-error value loss between terminal-return
and MC targets, and its RMS across $N$ responses, are
\begin{equation}
  \Delta g_i(A_i)=\frac{2}{|A_i|}\sum_{t\in A_i}
  \bigl(\widehat V_{\mathrm{MC}}^\pi(s_{i,t})-R_i\bigr)h_{i,t},
  \qquad
  D_{\mathrm{RMS}}=\sqrt{\frac{1}{N}\sum_{i=1}^N
  \|\Delta g_i(A_i)\|_2^2}.
  \label{eq:rms-value-head-gradient-discrepancy}
\end{equation}
\FloatBarrier

\subsection{Theoretical Analysis of Value Flattening}
\label{app:theoretical-insights}

We formalize two aspects of Value Flattening. We first decompose critic error
into response-mean and within-response components and examine how the latter
affects finite-batch actor updates. We then show that dense token-level
supervision introduces an implicit penalty on within-response prediction
variance and compare its gradient with a reference gradient based on exact
policy-conditioned state values. Finally, we explain why these finite-batch
effects do not conflict with the standard invariance of the expected policy
gradient to a state-dependent baseline.

\paragraph{Decomposing critic error.}
Let $i\in\{1,\ldots,N\}$ index responses and
$t\in\{1,\ldots,m_i\}$ index evaluated states.  Define the policy-conditioned
value $q_{i,t}=V^\pi(s_{i,t})$, the critic prediction
$v_{i,t}=V_\phi(s_{i,t})$, and, for any per-state quantity $x_{i,t}$,
write its response mean as
$\bar x_i:=m_i^{-1}\sum_{t=1}^{m_i}x_{i,t}$. In particular, this defines
$\bar q_i$ and $\bar v_i$.
The mean squared critic error, with each response weighted equally, decomposes as
\begin{align}
  \mathcal E_{\mathrm{point}}
  &:= \frac{1}{N}\sum_{i=1}^{N}\frac{1}{m_i}
      \sum_{t=1}^{m_i}(v_{i,t}-q_{i,t})^2 \\
  &= \underbrace{\frac{1}{N}\sum_{i=1}^{N}
      (\bar v_i-\bar q_i)^2}_{\mathcal E_{\mathrm{mean}}}
   + \underbrace{\frac{1}{N}\sum_{i=1}^{N}\frac{1}{m_i}
      \sum_{t=1}^{m_i}
      \left[(v_{i,t}-\bar v_i)-(q_{i,t}-\bar q_i)\right]^2}
      _{\mathcal E_{\mathrm{within}}}.
  \label{eq:value-resolution-decomposition}
\end{align}
The two terms measure error in the response mean and error in value changes
within the response, respectively. If states are
instead averaged over the entire batch, response $i$ receives weight
$\alpha_i=m_i/\sum_jm_j$, and the same decomposition holds.
Let
\begin{equation}
  e_{i,t}:=v_{i,t}-q_{i,t}
  =\bar e_i+e^\circ_{i,t},
  \qquad
  \frac{1}{m_i}\sum_{t=1}^{m_i}e^\circ_{i,t}=0,
  \label{eq:app-critic-error-decomposition}
\end{equation}
where $\bar e_i$ is the response-mean error and $e^\circ_{i,t}$ is the centered
within-response error.

\paragraph{From critic error to actor updates.}
Under $\gamma=\lambda=1$, all states in a response share the same realized
return. Let $R_i:=R(\tau_i)$ denote that terminal reward. The critic error
therefore changes the estimated advantage by $-e_{i,t}$, and its within-response
component changes the relative credit assigned to states by $-e^\circ_{i,t}$. Let
$z_{i,t}=\nabla_\theta\log\pi_\theta(a_{i,t}\mid s_{i,t})|_{\theta=\theta_{\mathrm{old}}}$
be the log-policy gradient and $\bar z_i=m_i^{-1}\sum_t z_{i,t}$ its response
mean. At the rollout policy, define the actor gradients obtained with the
sampled critic baseline and with a reference baseline using exact
policy-conditioned state values:
\begin{equation}
\begin{aligned}
  g_{\mathrm{critic}}
  &:= \frac{1}{N}\sum_{i=1}^{N}\frac{1}{m_i}
      \sum_{t=1}^{m_i}z_{i,t}\bigl(R_i-v_{i,t}\bigr),\\
  g_{\mathrm{oracle}}
  &:= \frac{1}{N}\sum_{i=1}^{N}\frac{1}{m_i}
      \sum_{t=1}^{m_i}z_{i,t}\bigl(R_i-q_{i,t}\bigr).
\end{aligned}
\label{eq:app-actor-gradients}
\end{equation}
We define the finite-batch actor-gradient error as
$\Delta g:=g_{\mathrm{critic}}-g_{\mathrm{oracle}}$. It then decomposes exactly as
\begin{equation}
  \Delta g
  = \underbrace{-\frac{1}{N}\sum_{i=1}^{N}
      \bar z_i\bar e_i}_{\Delta g_{\mathrm{mean}}}
    - \underbrace{\frac{1}{N}\sum_{i=1}^{N}\frac{1}{m_i}
      \sum_{t=1}^{m_i}(z_{i,t}-\bar z_i)e^\circ_{i,t}}_
      {\Delta g_{\mathrm{within}}}.
  \label{eq:app-gradient-error-decomposition}
\end{equation}
The first term is caused by error in the response mean.
The second is caused by error in the relative value profile within a response.
By Cauchy--Schwarz,
\begin{align}
  \|\Delta g_{\mathrm{mean}}\|
  &\le
  \left(\frac{1}{N}\sum_i\|\bar z_i\|^2\right)^{1/2}
  \mathcal E_{\mathrm{mean}}^{1/2},
  \nonumber\\
  \|\Delta g_{\mathrm{within}}\|
  &\le
  \left(
    \frac{1}{N}\sum_i\frac{1}{m_i}
    \sum_t\|z_{i,t}-\bar z_i\|^2
  \right)^{1/2}
  \mathcal E_{\mathrm{within}}^{1/2}.
  \label{eq:app-gradient-error-bounds}
\end{align}
Thus, each gradient-error component is bounded by the critic RMSE multiplied by
the RMS norm of the corresponding log-policy gradient. These are norm bounds;
they do not by themselves assert that the error is determined by directional
alignment.

\paragraph{Implicit Variance Penalty in Dense Critic Supervision.}
In the terminal-only setting, consider the critic loss under dense token-level
supervision, with each response weighted equally and using the sampled target
$R_i$. Let
$J_{i,t}=\nabla_\phi V_\phi(s_{i,t})$ and
$\bar J_i=m_i^{-1}\sum_tJ_{i,t}$ denote the critic Jacobian and its
response mean. The loss can then be decomposed as
\begin{equation}
  \mathcal L_{V,\mathrm{dense}}^{\mathrm{sample}}(\phi)
  := \frac{1}{N}\sum_{i=1}^{N}\frac{1}{m_i}
      \sum_{t=1}^{m_i}\bigl(v_{i,t}-R_i\bigr)^2
  = \frac{1}{N}\sum_{i=1}^{N}\left[
      (\bar v_i-R_i)^2
      +\frac{1}{m_i}\sum_{t=1}^{m_i}(v_{i,t}-\bar v_i)^2
    \right].
  \label{eq:app-dense-critic-loss}
\end{equation}
Its gradient is correspondingly
\begin{equation}
  \nabla_\phi\mathcal L_{V,\mathrm{dense}}^{\mathrm{sample}}
  = \frac{2}{N}\sum_{i=1}^{N}(\bar v_i-R_i)\bar J_i
    +\frac{2}{N}\sum_{i=1}^{N}\frac{1}{m_i}
      \sum_{t=1}^{m_i}(v_{i,t}-\bar v_i)(J_{i,t}-\bar J_i).
  \label{eq:app-dense-critic-gradient}
\end{equation}
The first term fits the mean critic prediction for each response to its sampled
terminal return. The second is the gradient of the within-response
prediction-variance term
$m_i^{-1}\sum_t(v_{i,t}-\bar v_i)^2$; the corresponding loss term directly
penalizes deviations from the response mean and therefore creates a flattening
pressure within each response.
The shared return provides no direct information about value differences among
states within a response, yet applying it at every state introduces this
variance penalty. The same
decomposition holds for token-weighted training, with response weights
$\alpha_i$.

To compare the PPO update in \cref{eq:app-dense-critic-gradient} with the update
needed to recover the policy-conditioned value profile, consider a hypothetical
regression to the exact values $q_{i,t}=V^\pi(s_{i,t})$. The corresponding gradients
of \cref{eq:value-resolution-decomposition} are
\begin{align}
  \nabla_\phi \mathcal E_{\mathrm{mean}}
  &= \frac{2}{N}\sum_{i=1}^{N}\bar e_i\bar J_i,
  &
  \nabla_\phi \mathcal E_{\mathrm{within}}
  &= \frac{2}{N}\sum_{i=1}^{N}\frac{1}{m_i}
      \sum_{t=1}^{m_i}e^\circ_{i,t}(J_{i,t}-\bar J_i)
    = \frac{2}{N}\sum_{i=1}^{N}\frac{1}{m_i}
      \sum_{t=1}^{m_i}e^\circ_{i,t}J_{i,t}.
  \label{eq:app-value-resolution-gradients}
\end{align}
Unlike the sampled-target update above, these are reference gradients for this
exact-value regression. Because
$m_i^{-1}\sum_t e^\circ_{i,t}=0$, the response-shared Jacobian component
cancels exactly. In other words, correcting within-response error requires the
state-level Jacobians to differ across states.

\paragraph{Expected-gradient invariance and practical scope.}
For any action-independent baseline $b(s_t)$,
\begin{equation}
  \mathbb E_{a_t\sim\pi(\cdot\mid s_t)}
  \left[
    \nabla_\theta\log\pi_\theta(a_t\mid s_t)b(s_t)
    \,\middle|\,s_t
  \right]
  =
  b(s_t)\nabla_\theta\sum_{a_t}\pi_\theta(a_t\mid s_t)
  =0.
  \label{eq:app-baseline-invariance}
\end{equation}
Exact pointwise values are therefore unnecessary for preserving the expected
on-policy gradient. On a finite batch, however, centered critic errors can have
a nonzero batch average when multiplied by centered log-policy gradients.

The finite-batch decomposition in \cref{eq:app-gradient-error-decomposition}
is exact for the local gradient at the rollout policy. Across multiple PPO
epochs, value errors can also alter advantage signs, magnitudes, and which
branch of PPO's clipped objective is active, so baseline invariance does not
imply identical clipped updates.

In summary, our analysis yields three observations:
\begin{itemize}[leftmargin=*,nosep]
  \item Reusing the same terminal return at every state introduces an implicit
    variance penalty on critic predictions within each response.
  \item Within-response critic error can change finite-batch actor updates
    through its interaction with centered policy gradients.
  \item Expected policy-gradient invariance does not imply identical practical
    PPO updates on finite batches.
\end{itemize}
Together, these results motivate evaluating whether critic predictions track
state-value changes within responses, in addition to response-level accuracy.
Combined with the temporal-correlation evidence in
\cref{sec:value-coarsening-analysis}, they motivate sparse critic supervision,
which applies the shared terminal return to fewer, well-separated states.

\subsection{Additional Comparisons of Critic Predictions and MC Values}
\label{app:value-resolution-recovery-cases}

\begin{figure}[H]
  \centering
  \includegraphics[width=0.98\linewidth]{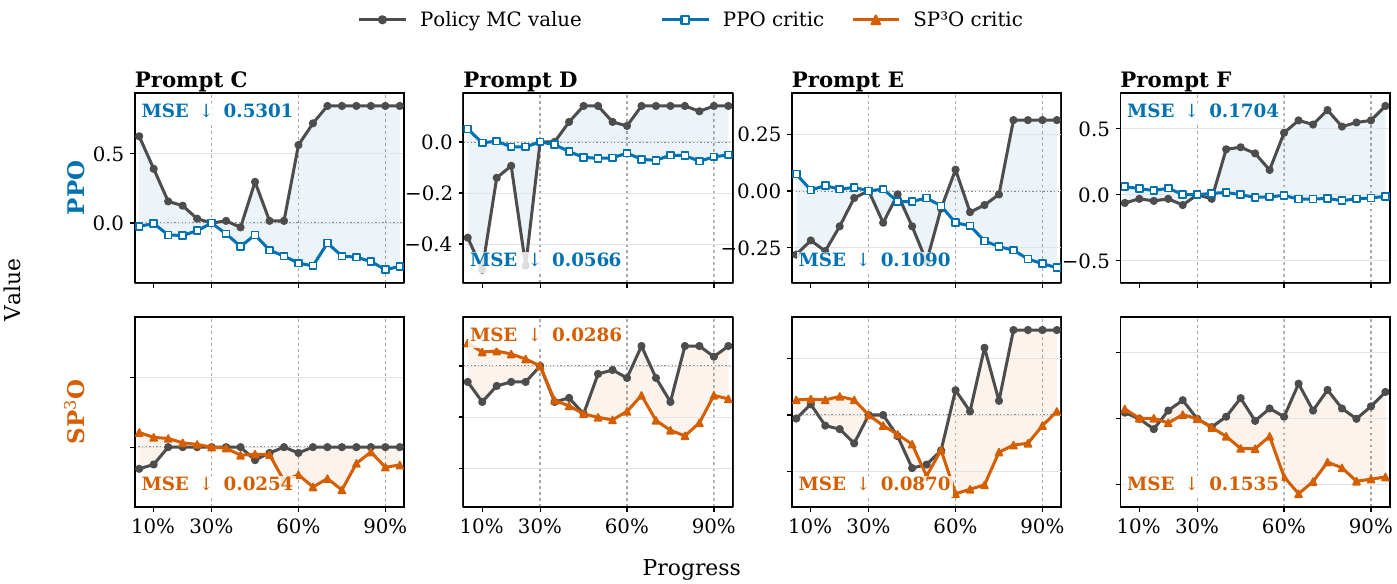}
  \caption{Additional prompt-matched value profiles. The horizontal axis is
  normalized response progress, and the vertical axis shows response-centered
  own-policy MC values and critic predictions.}
  \label{fig:value-resolution-recovery-additional-cases}
\end{figure}

\Cref{fig:value-resolution-recovery-additional-cases} complements the two
prompt-matched examples in \cref{fig:value-resolution-recovery-cases} with
four additional prompts from the same checkpoint evaluation panel. For
these prompts, \methodname{} has lower centered profile MSE than PPO.

\FloatBarrier
\subsection{Sparse-Supervision Ablations}
\label{app:sparse-supervision-ablations}
\label{app:anchor-position-ablation}

\begin{table}[!ht]
    \centering
  \footnotesize
  \setlength{\tabcolsep}{5pt}
  \begin{tabular}{@{}lrr@{}}
    \toprule
    \textbf{Variant} & \textbf{Acc. (\%)}
      & \textbf{Repetition (\%)} \\
    \midrule
    \methodname{} w/o tail anchor & 44.10 & 18.33 \\
    \methodname{}    & \textbf{45.57} & \textbf{1.12} \\
    \bottomrule
  \end{tabular}
  \caption{Late-tail ablation for \methodname{}.}
  \label{tab:anchor-ablation-tail}

\end{table}

\begin{wrapfigure}{r}{0.46\linewidth}
  \vspace{-0.8\baselineskip}
  \centering
  \includegraphics[width=0.92\linewidth]{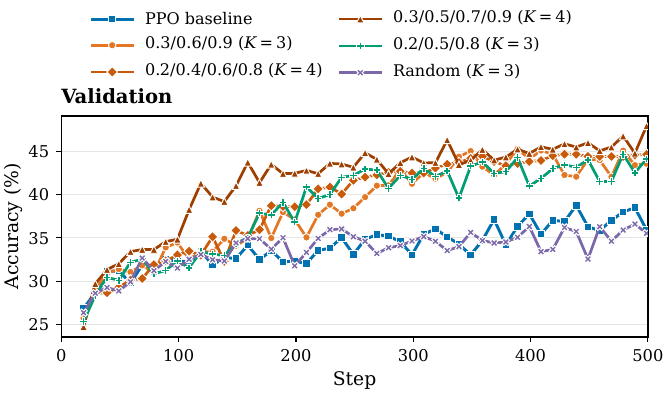}
  \caption{Performance across critic supervision anchor placements.}
  \label{fig:anchor-position-ablation-curves}
  \label{fig:anchor-ablation-figures}
  \vspace{-0.7\baselineskip}
\end{wrapfigure}

We examine how the number and placement of critic-supervision anchors affect
Qwen3-4B-Base. The placement curves show that later-state coverage is generally
beneficial, whereas adding anchors alone does not reliably improve performance.
The matched late-tail comparison is reported in \cref{tab:anchor-ablation-tail}.
Conditional tail coverage provides a further improvement and reduces repetitive behavior.
\Cref{fig:anchor-ablation-figures} shows the corresponding online validation
trajectories. Schedules with later-state coverage achieve higher validation
accuracy, consistent with the placement comparison in \cref{fig:anchor-position-ablation-curves}.

\FloatBarrier

\end{document}